\documentclass{article}
\usepackage{microtype}
\usepackage{graphicx}
\usepackage{subfigure}
\usepackage{booktabs} % for professional tables
\usepackage{adjustbox}
\usepackage{hyperref}

\usepackage[accepted]{mlsys2022}

\graphicspath{ {./images/} }

\begin{document}

\twocolumn[
\mlsystitle{Video-Data Pipelines for Machine Learning Applications}



\begin{mlsysauthorlist}
\mlsysauthor{Sohini Roychowdhury}{to}
\mlsysauthor{James Y. Sato }{too}
\end{mlsysauthorlist}

\mlsysaffiliation{to}{Director Curriculum, Fourthbrain, CA-95050}
\mlsysaffiliation{too}{FourthBrain, CA-95050}

\mlsyscorrespondingauthor{Sohini Roychowdhury}{roych@uw.edu}
\mlsyskeywords{Machine Learning, Production, Data pipeline, video-sequences, object detection}

\vskip 0.3in

\begin{abstract}
Data pipelines are an essential component for end-to-end solutions that take machine learning algorithms to production. Engineering data pipelines for video-sequences poses several challenges including isolation of key-frames from video sequences that are high quality and represent significant variations in the scene. Manual isolation of such quality key-frames can take hours of sifting through hours’ worth of video data. In this work, we present a data pipeline framework that can automate this process of manual frame sifting in video sequences by controlling the fraction of frames that can be removed based on image quality and content type. Additionally, the frames that are retained can be automatically tagged per sequence, thereby simplifying the process of automated data retrieval for future ML model deployments. We analyze the performance of the proposed video-data pipeline for versioned deployment and monitoring for object detection algorithms that are trained on outdoor autonomous driving video sequences. The proposed video-data pipeline can retain anywhere between 0.1-20\% of the all input frames that are representative of high image quality and high variations in content. This frame selection, automated scene tagging followed by model verification can be completed in under 30 seconds for 22 video-sequences under analysis in this work. Thus, the proposed framework can be scaled to additional video-sequence data sets for automating ML versioned deployments.
\end{abstract}
]

\section{Introduction}
Computer Vision applications rely of large volumes of video data to be processed to learn the patterns related to the objects/regions of interest (ROIs). From robotic vision to object detection and real-time object tracking applications for autonomous driving, there is a need to isolate \textit{key-frames} with \textit{high quality} from long sequence of videos that can then train respective machine learning (ML) applications \cite{videochallenges}. In this work, we present a video-data processing pipeline that can be combined with modeling and deployment pipelines specifically for video/image-based ML use-cases. 

In this work we focus on autonomous-driving video sequences captured using a 1.4 megapixel dashboard mounted camera in a moving vehicle. These video frames correspond to high quality outdoor images that may suffer from glare, low standing sun, shadow contrasts, but represent no major artifacts or motion blur. Such a data-pipeline can be easily modified to accommodate indoor and handheld video sequences that may present additional motion-based artifacts.

\begin{figure}[ht]
\includegraphics[width=\columnwidth,keepaspectratio]{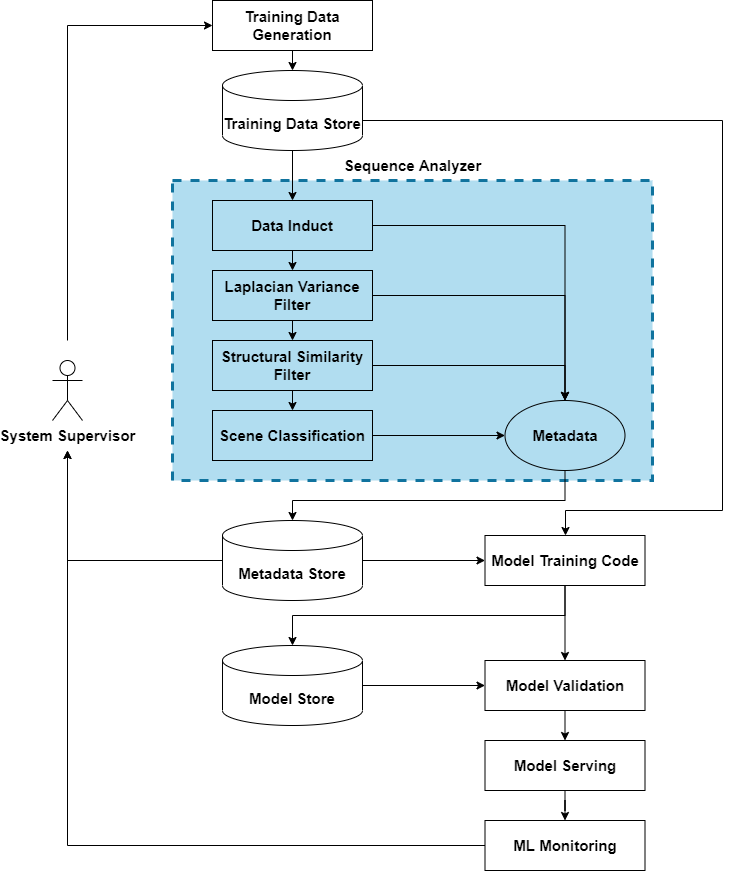} 
\caption{Proposed workflow for Video-data Pipelines applied to ML solutions in Production. The highlighted region represents the proposed video-data pipeline that feeds into the ML modeling, serving/deployment and monitoring pipelines for versioned ML deployments with minimal manual supervision.}\label{system}
\end{figure}

This paper makes two major contributions. First, we present a data pipeline to pre-process and retain a fraction of sequential frames by analyzing per frame quality and content-type. These retained frames can then used for subsequent training of ML models. The frame selection process is followed by automated metadata tagging for the selected/retained frames, which enables automated storage and retrieval of frames per sequence. Second, we present ML modeling, deployment and monitoring pipelines that utilize the proposed video-data pipeline for versioned model deployments and that can be scaled and modified based on use-case and business needs. The proposed workflow for end-to-end delivery of ML solutions in production is shown in Fig. \ref{system}.

\section{Related Work}
To deploy ML solutions for production and scaling, a variety of MLOps pipelines \cite{mlops} have been introduced wherein the data, modeling, and deployment and monitoring pipelines can be automated at a variety of levels to suit the application and business needs.

Several works so far have outlined the need for modeling and unit testing in ML workflows such as in \cite{breck2018data} and \cite{breck2019data}. Additionally, the work in \cite{breck2019data} presents a schema to detect errors in data pipelines for ML solutions. However, video-based data processing pipelines pose additional challenges related to \textit{key-frame} identification \cite{keyframe} and efficient data storage and retrieval \cite{videochallenges}. In this work, we present end-to-end ML pipelines that are capable of ingesting large volumes of video data streams, identifying the key-frames that need annotation, and creating an efficient storage-retrieval schema that aid data collection for subsequent modeling cycles. 

One well-known \textit{Big Data Solution} for data pipelines includes construction of data lakes with the Hadoop Distributed File System \cite{adbench}. However, since annotation costs impact video-based ML solutions very heavily, there is a need to streamline the process of frame identification, and to curate sequences and frame-by-frame metadata for faster storage and retrieval operations. In this work, we present a data pipeline that extracts key-frames from outdoor video sequences that can then be stored using additional metadata based on the automatically detected contents of the frames. We demonstrate the plug-and-play nature of the proposed pipeline with modeling and deployment pipelines for end-to-end delivery of computer-vision solutions. 

\section{Data and Methods}
In this work, we propose a compute-optimized video-data pipeline that supports deep learning models \cite{videochallenges} for classification, object detection, and semantic segmentation tasks. Training deep learning models require large volumes of image frames that are of high image quality and that significantly vary in lighting conditions and image content to ensure that the trained deep learning model generalizes the patterns for predicting ROIs on new test data sets.

\subsection{Data}
In this work, we develop an automated data pre-processing pipeline for video sequences. We analyze the impact of the data processing modules on two public data sets to benchmark. The first KITTI tracking dataset \cite{KITTItracking} for 2D objects has a variety of video-sequences of outdoor scenes of varying lengths as shown in Table \ref{tab1}. The video sequences are acquired at a frame-rate of 10 frames-per-second and the images are cropped frames of [1382 x 512] pixels per frame. To further evaluate the utility of the proposed video-data pipeline, we analyze a stack of outdoor images from the 2D object detection benchmark data set in \cite{KITTItracking}, referenced as \textit{non-sequential} in Table \ref{tab1}.
\begin{table}[ht!]
\caption{Explanation of outdoor Video-datasets from \cite{KITTItracking}}
\begin{adjustbox}{width=0.5\columnwidth,center}
\begin{tabular}{|c|c|}
\hline
\textbf{Video} & \textbf{Frame} \\
\textbf{Sequence$^{\mathrm{1}}$} & \textbf{Count} \\ \hline
0000 & 154  \\ \hline
0001 & 447  \\ \hline
0002 & 233  \\ \hline
0003 & 144  \\ \hline
0004 & 314  \\ \hline
0005 & 297  \\ \hline
0006 & 270  \\ \hline
0007 & 800  \\ \hline
0008 & 390  \\ \hline
0009 & 803  \\ \hline
0010 & 294  \\ \hline
0011 & 373  \\ \hline
0012 &  78  \\ \hline
0013 & 340  \\ \hline
0014 & 106  \\ \hline
0015 & 376  \\ \hline
0016 & 209  \\ \hline
0017 & 145  \\ \hline
0018 & 339  \\ \hline
0019 & 1059 \\ \hline
0020 & 837  \\ \hline
non-sequential& 7481 \\ \hline
\end{tabular}
\label{tab1}
\end{adjustbox}
\end{table}

\subsection{Video-Data Pipeline Components}
To process the sequence of frames as described above, we propose a two-step image-data filtering technique that eliminates frames with inferior \textit{image quality} and images that contain objects and lighting conditions that have already been encountered in prior frames. For the proposed first step, we applied a Laplacian filter to detect images with limited foreground definition. These images represent \textit{blurry, poor illumination, artifacts}, and \textit{low standing sun} that can negatively impact ML model training. In the second step, structural similarly between consecutive video frames is assessed to eliminate frames with similar structural content compared to a previously selected frame. Finally, a pre-trained object detection model is applied to detect objects in each frame. The detected objects are aggregated using categorical aggregation to enable automated metadata curation and tagging for each retained frame. These components are described in the sections below.

\subsubsection{Frame Quality Analysis for Image-data Filtration}\label{filter}
Image blurriness, lighting conditions, and artifacts are some of the primary issues of image quality analysis. To isolate \textit{high quality frames}, we apply video-data filtration wherein, we eliminate image frames with low image clarity using the OpenCV Laplacian variance function based on existing works in \cite{blur} and \cite{blur2}. The 2D Laplacian kernel highlights foreground edges, thereby yielding a higher score for pixel variance in well-focused and sharp images. We apply the Laplacian function to grayscale images converted from the video sequences and retain high quality images with high coefficients for the modeling pipeline. 

We analyzed other metrics for image-quality selection such as Peak signal to noise ration, contrast to noise ratio, coefficient of variations, but none of the other metrics presented a wide parametric range as offered by the $VOL$ method. Additionally, we focus on simpler metrics over contrastive learning metrics \cite{SISE-PC} or deep-learning based metrics for this initial data selection phase to reduce computational time complexity of the data pipeline and promote its adoption for automated ML modeling and deployment pipelines.

The image quality analysis implementation steps are as follows. First, the Variance of Laplacian ($VOL$) per frame is evaluated in a synchronized pool across CPU cores to ensure the process scales with computation resources. Second, image frames with $VOL$ lesser than a threshold value $v$ (usually $v \in [300,1000]$) are filtered away from the sequence stack. Finally, a data dictionary is returned containing the following metadata: the per frame $VOL$, and its distribution statistics (minimum, maximum, and median $VOL$ for the stack), frame IDs, and number and ratio of frames removed from the video sequence.

Following the Laplacian filter, we further eliminate video frames using the structural similarity index metric ($SSIM$) based on the work in \cite{patent}. $SSIM$ captures the structural similarity between consecutive frames, thereby consecutive frames have $SSIM$ close to 1 while the $SSIM$ reduces as the image content changes significantly. The implementation of $SSIM$ to further filter frames goes as follows:

First, $SSIM$ per two frames in sequence is calculated in a synchronized pool. Second, frames with $SSIM$ greater than a threshold $s$ ($s \in [0, 1]$) are removed. Finally, the metadata dictionary from the previous step is appended with the fraction and IDs of frames that are removed from the video sequence using $SSIM$. 
\begin{figure*}[ht]
    \centering
	\subfigure[Image 0000/000145.png]
	{\includegraphics[width=0.45\textwidth]{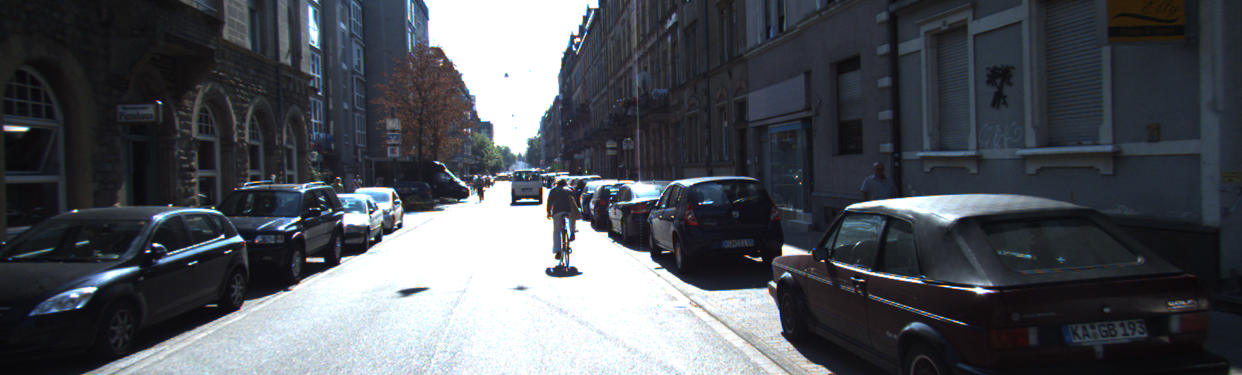}}
	\subfigure[Image 0004/000215.png]
	{\includegraphics[width=0.45\textwidth]{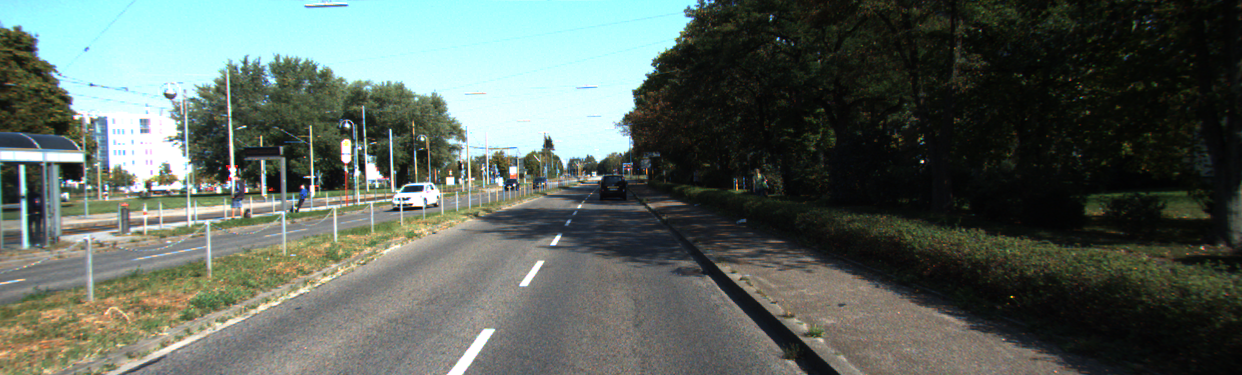}}
	\subfigure[Image 0007/000186.png]
	{\includegraphics[width=0.45\textwidth]{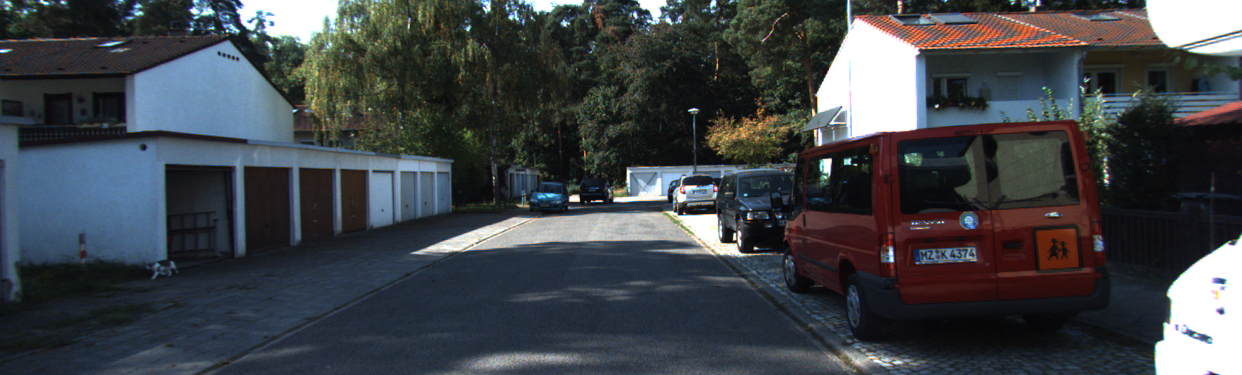}}
	\subfigure[Image 0008/000177.png]
	{\includegraphics[width=0.45\textwidth]{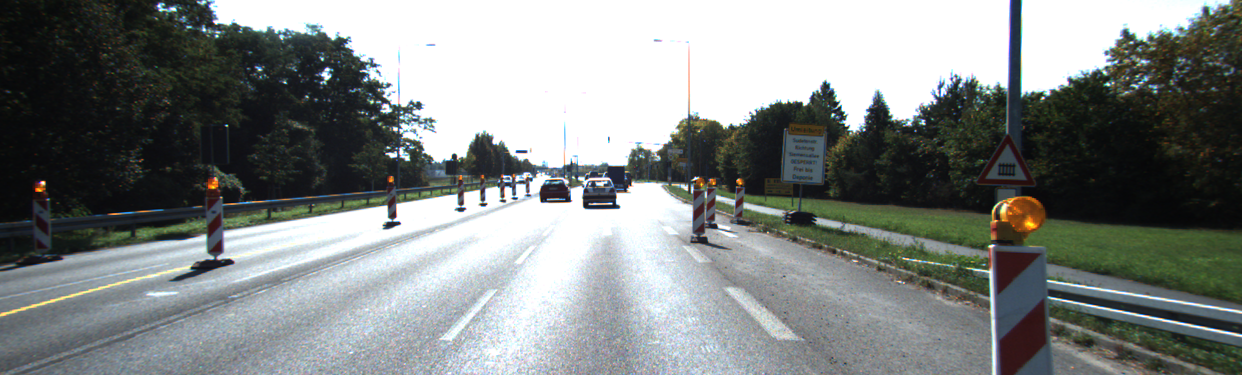}}
	\subfigure[Image 0009/000257.png]
	{\includegraphics[width=0.45\textwidth]{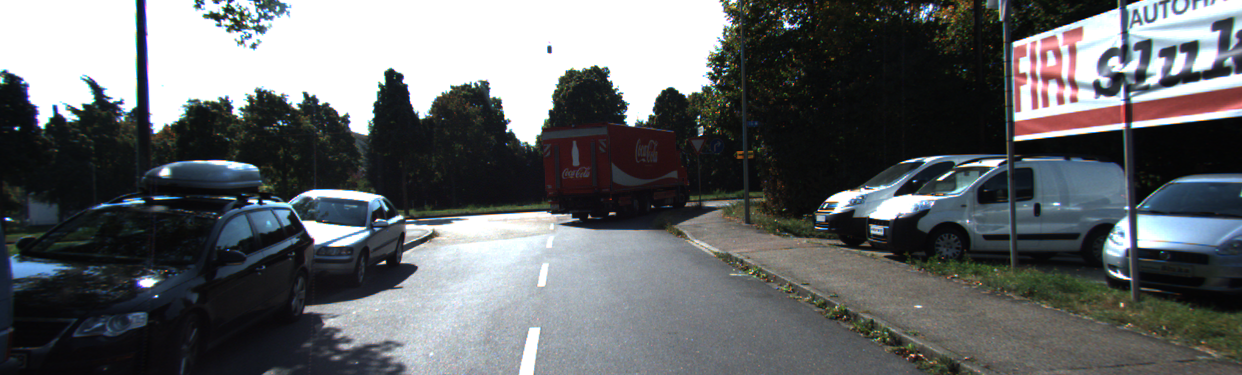}}
	\subfigure[Image 0010/000217.png]
	{\includegraphics[width=0.45\textwidth]{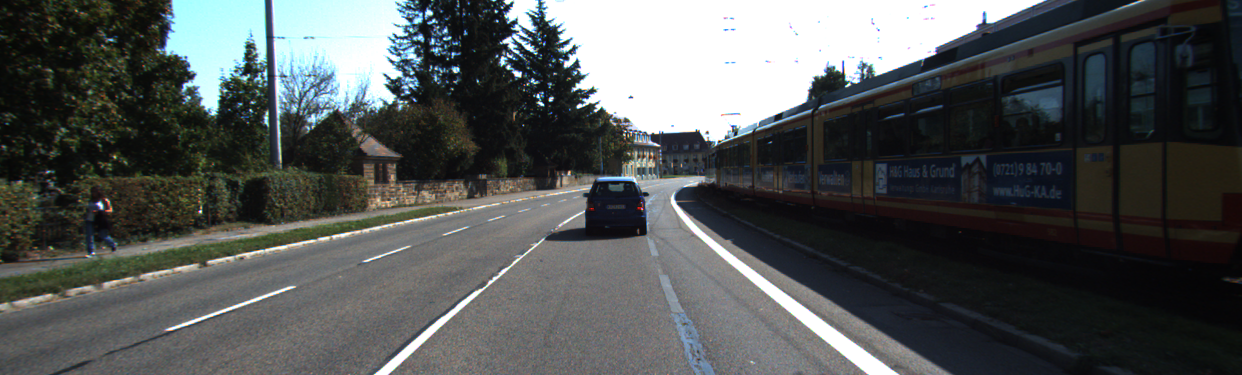}}
	\subfigure[Image 0011/000099.png]
	{\includegraphics[width=0.45\textwidth]{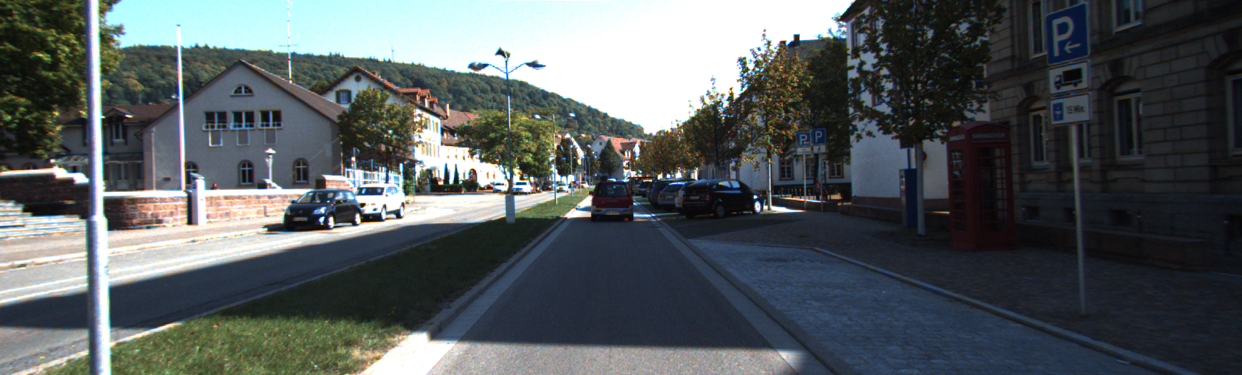}}
	\subfigure[Image 0019/000022.png]
	{\includegraphics[width=0.45\textwidth]{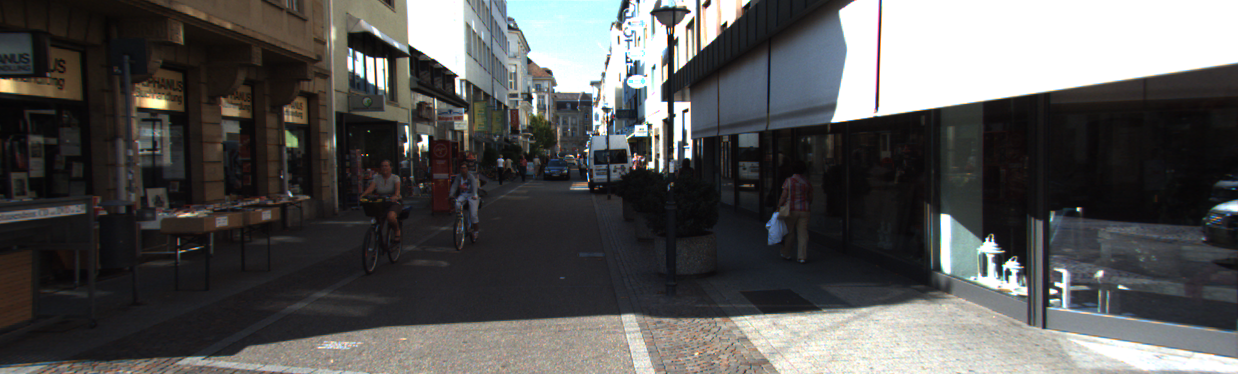}}
	\caption{Examples of Images filtered by the Sequence Analyzer}\label{ex1}
       \vspace{-0.3cm}
\end{figure*}

Some examples of images that are eliminated by the proposed two-step image quality filtering method are shown in Fig. \ref{ex1}. The sequence IDs and the frame IDs are provided in the sub-captions. Sub-images Fig. \ref{ex1} (b), (c), (e), (g), (h) are eliminated due to the shadow regions that result in low $VOL$ values per frame. The remaining images are eliminated based on their high $SSIM$ score since \textit{similar} images have already been retained previously in the video sequence.

All the images that are removed during the data filtering stage are considered as test data for the ML model that is created in the modeling and deployment phases of the workflow. The images that are retained in this step are to train/fine-tune the ML models in the next stages.

\subsubsection{Sequence Classification for Storage and Retrieval}
Upon isolating the training data set by data filtering from video sequences, the next task is to generate metadata to tag each frame and sequences to enable fast storage and retrieval for future ML model versions. For instance, if a specific ML model is currently being trained to aid an autopilot functionality using object detection in highways for cargo vehicles, then training images corresponding to highways and freeways will be needed to fine tune object detection models. However, for the next version of ML model that could specialize for pedestrian detection in city, the training images will need to be different. Here, auto-tagging of image frames enables quick retrieval of relevant images for such ML model versioning use-cases.

For the KITTI datasets under analysis in this work, a variety of objects labels are pre-annotated per frame. To aid scene classification, we first group combinations of objects as follows. The objects groups and the object labels corresponding to the object group are shown below.
\begin{itemize}
    \item Vehicles: \{`Car', `Van', `Truck'\}
    \item People: \{`Pedestrian', `Person\_sitting', `Cyclist'\}
    \item Urban Vehicle: \{`Tram'\}
\end{itemize}

Next, to classify a scene we apply the following rules based on the objects groups detected per frame. These rule-based scene categorizations are empirically generated by us and may be subjected to inter-observer variability. The rule-based scene categories and their corresponding objects groups are applied as shown below.
\begin{itemize}
    \item City: Scene with at least one `Urban Vehicle' or a combination of at least one `Vehicles' and/or one `People' object groups, respectively.
    \item Pedestrian Traffic: Scene with one or more `People' but no `Vehicles' or `Urban Vehicle'\\
    \item Freeway: Scene with at least 2 `Vehicles' but no `People' or `Urban Vehicle'\\
    \item Rural: Scene with at least one `Vehicle' and up to 2 `People'\\
    \item Unknown: Scene that does not fall under any of the other categories.
\end{itemize}
This grouping of objects and scene classification enables quick data searches for versioned model deployments that will be shown in the next sections. The code for creating the proposed video-data pipelines is available for bench-marking purposes. \footnote{The github repository for the code is at https://github.com/James-Yuichi-Sato/MLSYS-2022-Video-Pipeline}.

\subsection{ML Modeling Pipelines}
To assess the importance of the proposed video-data pipeline, we implement a ML model pipeline to utilize the auto-tagged data. The ML model pipeline includes searching for data batches related to the use-case, e.g. querying for `Freeway' and `City' scenes to aid highway autopilot functionalities. At the end of the ML model pipeline, the trained models are deployed and subjected to real-time monitoring that enables automated online supervision of the deployed ML model performance. Further, the modeling and deployment pipelines enable multiple versioned ML deployments based on the input data distributions or varying use-cases. 

For our ML modeling pipeline, We apply the images isolated from the video-data pipeline to fine-tune two versions of object detection algorithms, namely the FasterRCNN model \cite{faster} for bounding box detection and the Mask-RCNN \cite{mask} for more accurate bounding box detections and semantic segmentations. To train both the ML models, we apply transfer learning to fine-tune the respective pre-trained bounding box detection models from the Tensorflow hub \cite{TFhub}. We apply data augmentations, the Adam optimizer with learning rate of $10^{-3}$ and fine-tune the object detector model for 50 epochs with a batch size of 32 images per epoch. Upon training, the inference time per test frame, and the bounding box detection performance is analyzed to verify efficacy of the deep learning model. We detect all objects with probability of objectness greater than 0.1. The inference time per test frame is found to range between [0.15-0.17] seconds using the FasterRCNN model. 

\begin{figure*}[ht]
    \centering
	\subfigure[Image 0000/000145.png]
	{\includegraphics[width=0.45\textwidth]{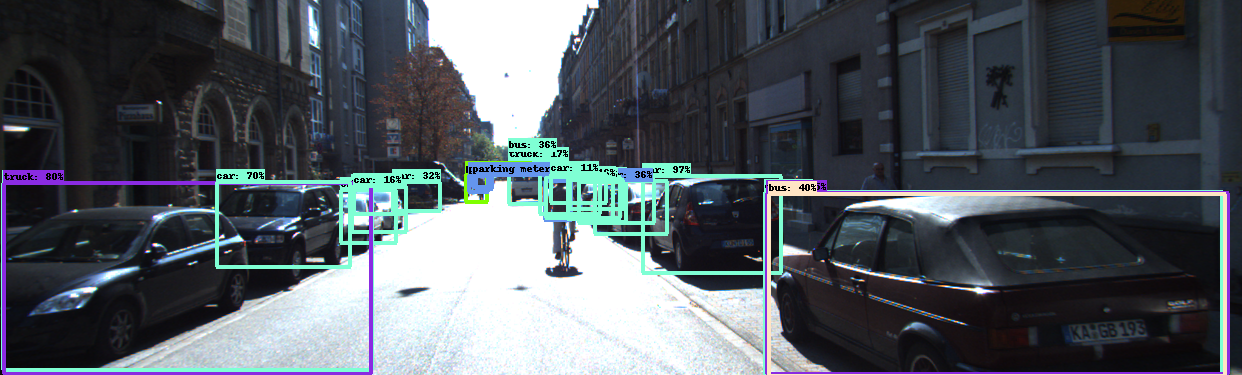}}
	\subfigure[Image 0004/000215.png]
	{\includegraphics[width=0.45\textwidth]{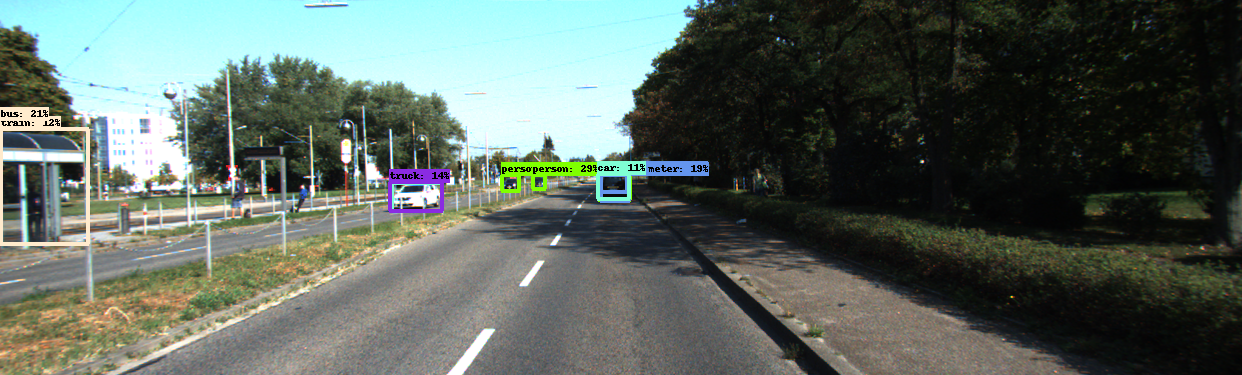}}
	\subfigure[Image 0007/000186.png]
	{\includegraphics[width=0.45\textwidth]{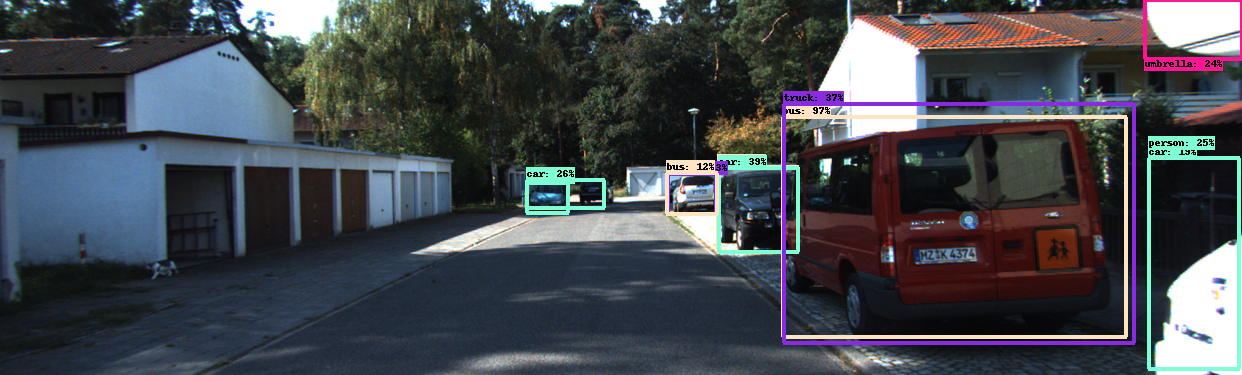}}
	\subfigure[Image 0008/000177.png]
	{\includegraphics[width=0.45\textwidth]{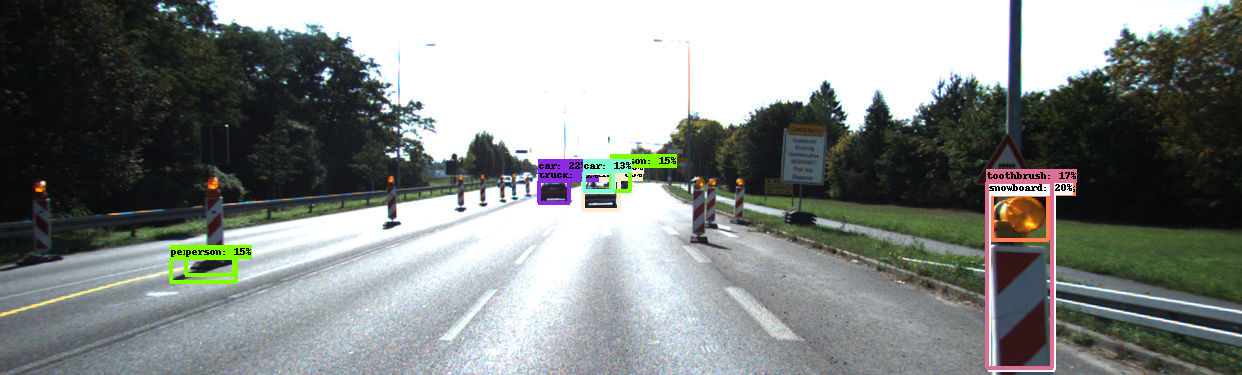}}
	\subfigure[Image 0009/000257.png]
	{\includegraphics[width=0.45\textwidth]{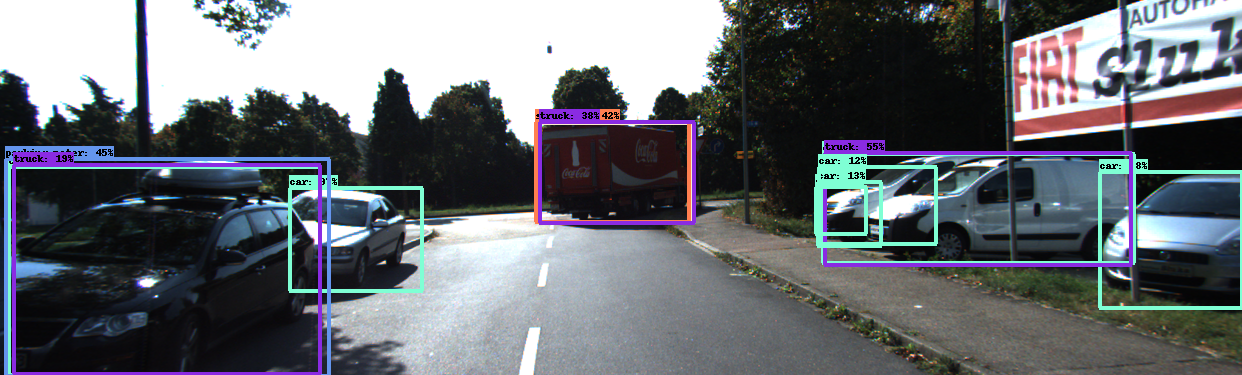}}
	\subfigure[Image 0010/000217.png]
	{\includegraphics[width=0.45\textwidth]{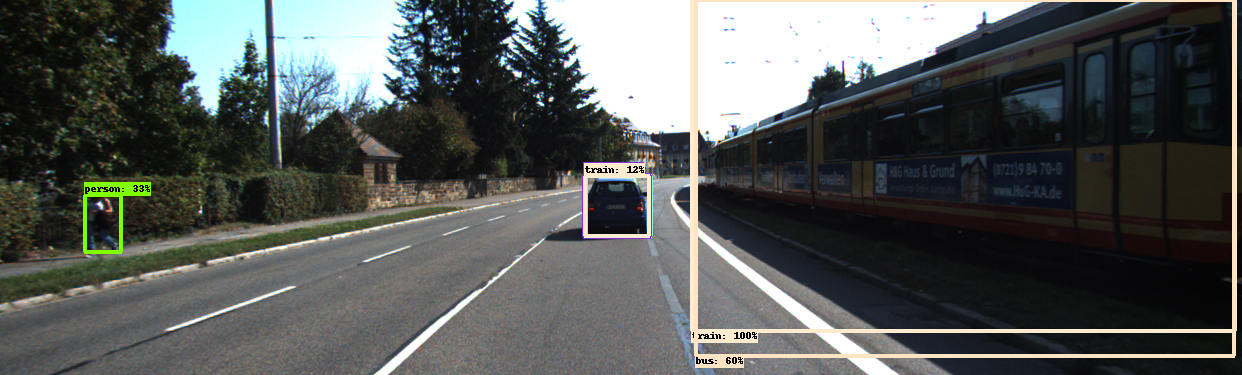}}
	\subfigure[Image 0011/000099.png]
	{\includegraphics[width=0.45\textwidth]{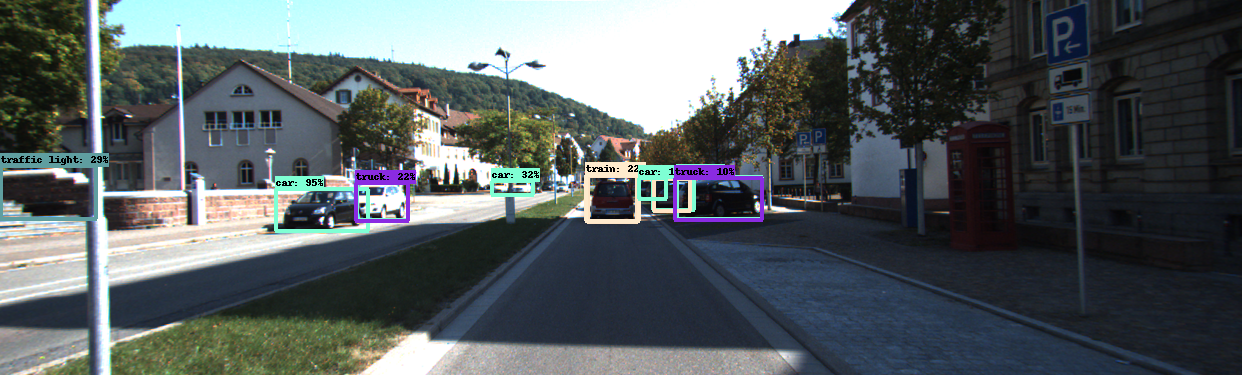}}
	\subfigure[Image 0019/000022.png]
	{\includegraphics[width=0.45\textwidth]{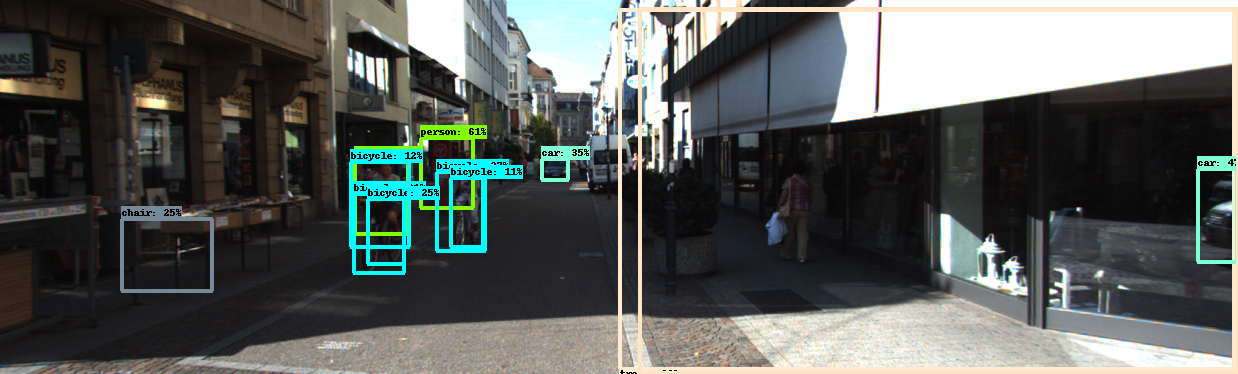}}
	\caption{Examples of Bounding-box detection using FasterRCNN model on the Filtered Frames.}\label{ex2}
       \vspace{-0.3cm}
\end{figure*}

Examples of the FasterRCNN model detections on the filtered/test images from Fig. \ref{ex1} are shown in Fig. \ref{ex2}. Here, we observe that most detections with lower probabilities (0.1-0.3) have multiple spurious class labels. However, the lower objectness threshold of 0.1 enables faraway object detection, which is a requirement for accurate scene classification. For our application of automated scene categorization, these spurious labels do not alter the scene classification operation, since a false classification label of `toothbrush' as shown in Fig. \ref{ex2}(d) does not fall under any object category of interest. 

Our next step is to enable versioned deployment of a more accurate object detection model over time that relies on semantic segmentation for predicting the probability of \textit{objectness} more accurately. For this purpose, we train and fine-tune a Mask-RCNN model \cite{mask} using the filtered images from Section \ref{filter}. The training process is continued for 50 epochs, Adam optimizer with learning rate $10^{-5}$ and objectness threshold of 0.1. For a trained MaskRCNN model, the inference time per test frame ranges from [0.65-0.71] seconds per frame.

\begin{figure*}[ht]
    \centering
	\subfigure[Image 0000/000145.png]
	{\includegraphics[width=0.45\textwidth]{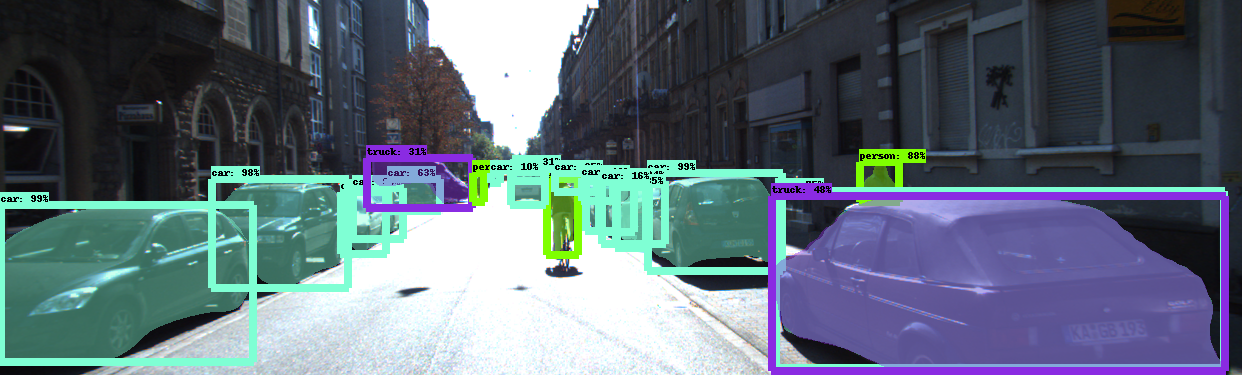}}
	\subfigure[Image 0004/000215.png]
	{\includegraphics[width=0.45\textwidth]{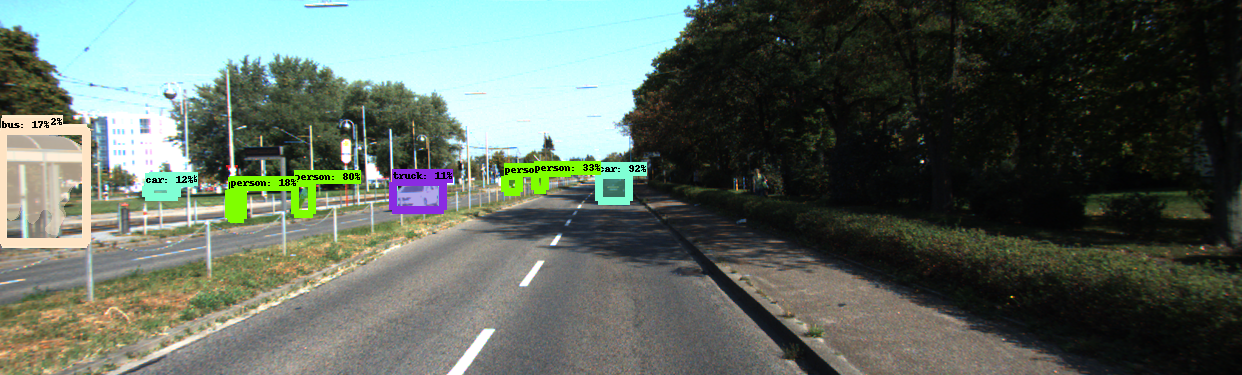}}
	\subfigure[Image 0007/000186.png]
	{\includegraphics[width=0.45\textwidth]{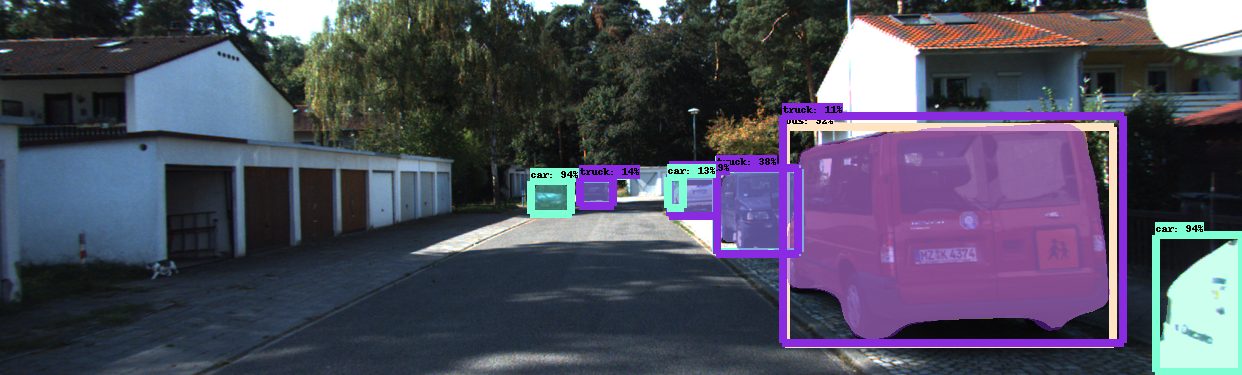}}
	\subfigure[Image 0008/000177.png]
	{\includegraphics[width=0.45\textwidth]{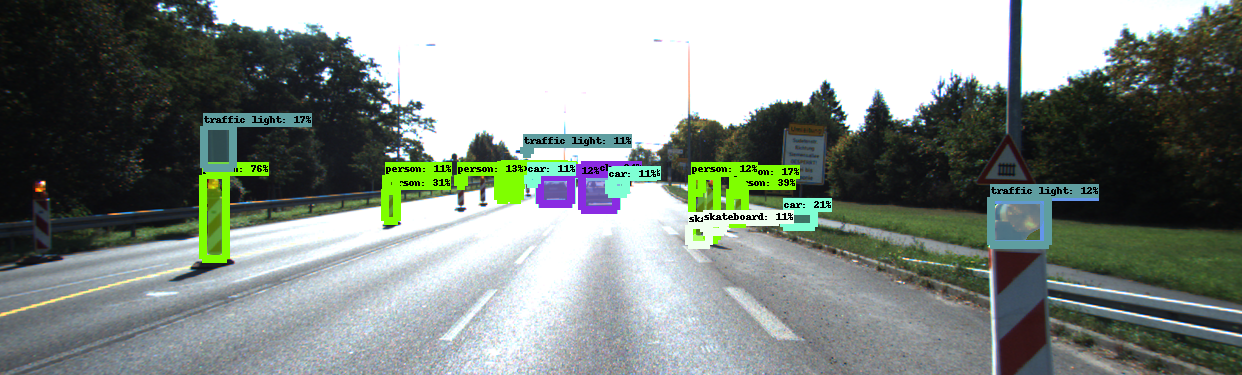}}
	\subfigure[Image 0009/000257.png]
	{\includegraphics[width=0.45\textwidth]{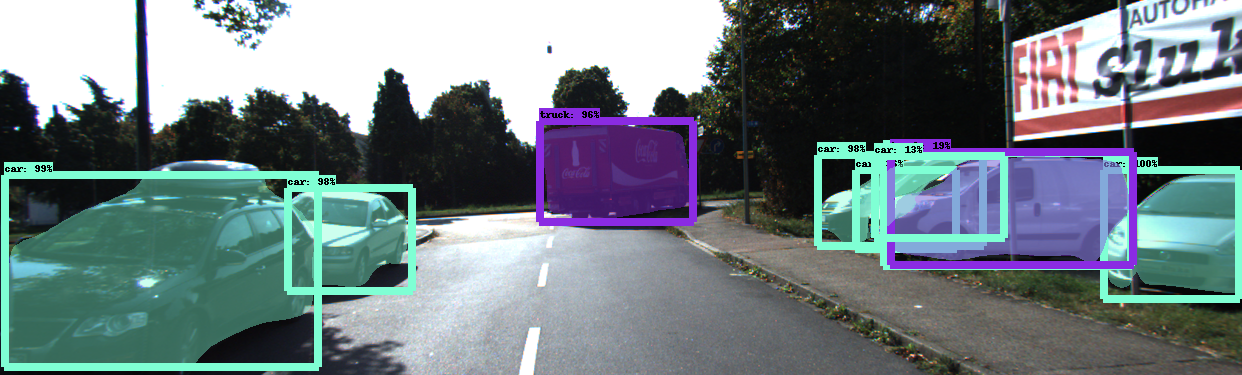}}
	\subfigure[Image 0010/000217.png]
	{\includegraphics[width=0.45\textwidth]{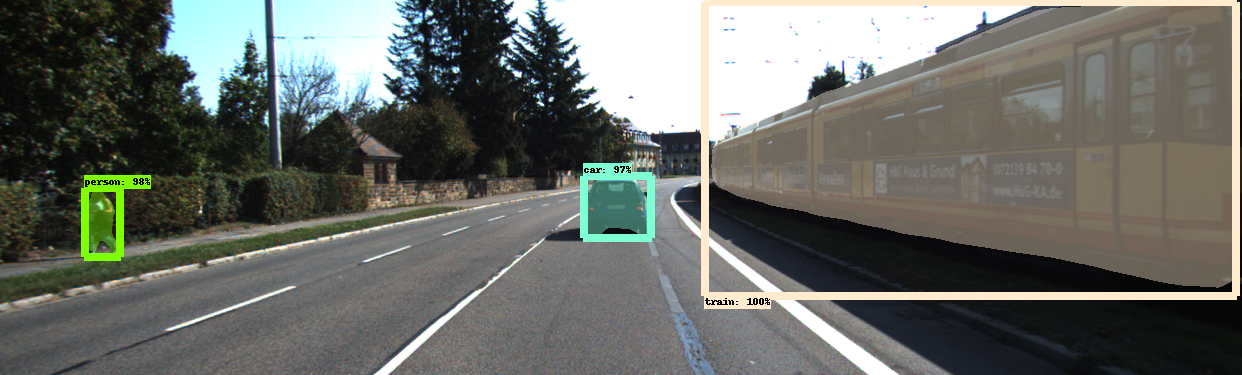}}
	\subfigure[Image 0011/000099.png]
	{\includegraphics[width=0.45\textwidth]{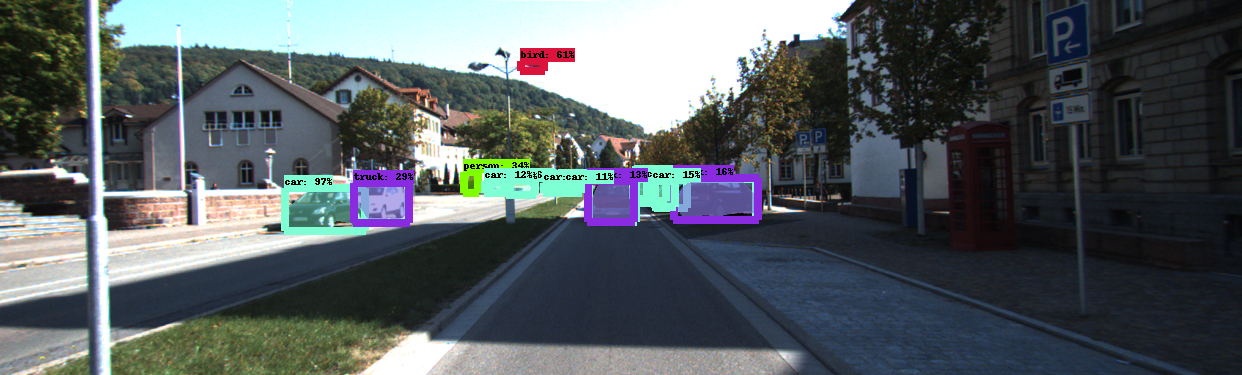}}
	\subfigure[Image 0019/000022.png]
	{\includegraphics[width=0.45\textwidth]{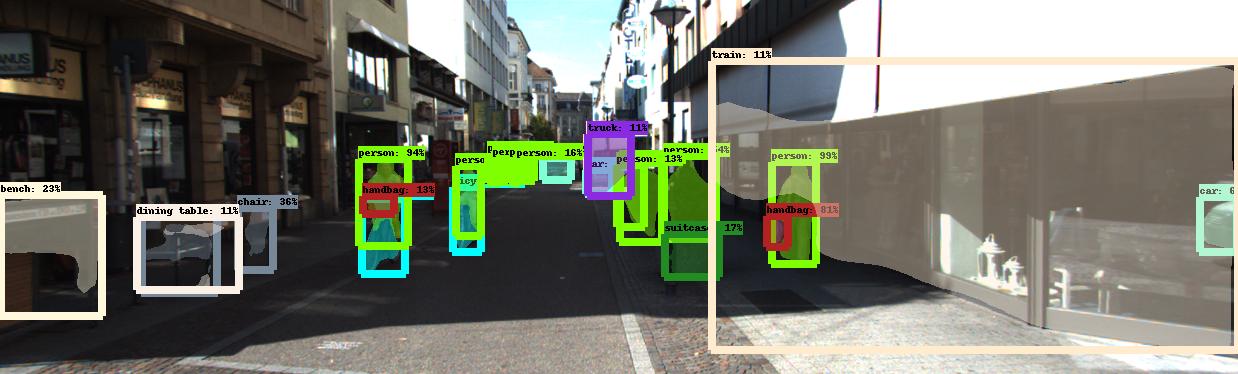}}
	\caption{Examples of Bounding-box and semantic segmentations using Mask-RCNN on the Filtered Frames.}\label{ex3}
       \vspace{-0.3cm}
\end{figure*}

Examples of Mask-RCNN model applied to the test images from Fig. \ref{ex1} are shown in Fig. \ref{ex3}. We observe more objects detected by Mask-RCNN in Fig. \ref{ex3} than by FasterRCNN in Fig. \ref{ex2}. Semantic segmentation in Mask-RCNN further suppresses spurious object labels such as in Fig. \ref{ex3}(d) over Fig. \ref{ex2}(d). Thus, a trained Mask-RCNN model can serve as an object detection model upgrade to FasterRCNN, but with a slower processing time. Thus offline object detection use-cases can be better served by Mask-RCNN over FasterRCNN.

\subsection{Deployment and Monitoring Pipelines}
\begin{figure}[ht]
     \centering
    \includegraphics[width=3.3in, keepaspectratio=True]{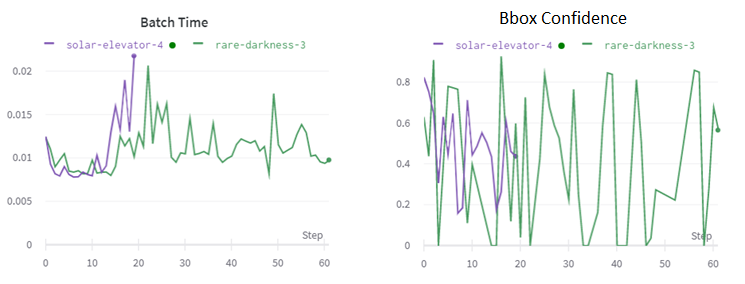}
        \caption{Example of ML monitoring for outlier detection in test data.}
    \label{grafana}
    \vspace{-0.3cm}
\end{figure}

Once an ML model is trained, it needs to be deployed, scaled and monitored using online performance metrics. For our application, we dockerize the ML models and deploy them over a Kubernetes cluster that can automatically scale to produce 2-20 replica pods based on usage \cite{deploy}. Upon deployment, the next task is to monitor the performance of the ML model on live incoming data streams. Here the online performance metrics we use are the processing times per batch of frames and mean bounding box confidence probability (average probability of objectness) per frame as shown in Fig. \ref{grafana}. Additionally, ML model and Kubernetes cluster health monitoring dashboards can be setup using the Prometheus gateway and Grafana dashboards \cite{promgraph}.

ML model monitoring enables model serving/deployment tracking and scaling based on concept and data drifts for incoming test batches of data. In the event when the monitoring performance metrics show significant changes, such as an increase in processing time and/or significant reduction in mean bounding box probabilities per frame batch, an improved ML model version can be rolled out using shadow, canary or blue-green deployment patterns over the Kubernetes cluster with updated fine-tuned ML models \cite{canary}. Thus, the proposed video-data pipeline aids for quick retrieval of this additional data for versioned deployments. For our application, we analyze the impact of the proposed video-data pipeline to roll-out a first version of object detection ML model using FasterRCNN, followed by a canary roll-out of a fine-tuned Mask-RCNN deployment.

\section{Experiments and Results}
We analyze the performance of the proposed video-data pipeline under various settings and parametric constraints to gauge its generalizability across use cases. Here, we perform two major experiments. First, we assess the thresholds for the $VOL$ and $SSIM$ filters to vary the ratio of filtered vs. unfiltered frames. Second, we assess the scene categorization performance to verify the usefulness of automated frame tagging.The experiments and observations are described in the following subsections.

\subsection{Thresholding the $VOL$ Filter}
As the first step, the $VOL$ filter has a significant impact on the volumes of unfiltered/retained frames that must subsequently get annotated to train a ML model. We observe a wide variability in $VOL$ values across sequences. We empirically gauge that for outdoor sequences generated for autonomous driving, such as the KITTI data set, a $VOL$ threshold of 500 is generally low, such that retaining frames with $VOL>500$ results in retaining most frames in the sequences. However, we observe that retaining frames over $VOL$ threshold values of 700 and 800, results in about 40\% and 25\% of the frames, respectively, to be retained as \textit{high quality}. Table \ref{tab3} shows the percentage of frames retained by varying the $VOL$ thresholds. This analysis aids redesigning the filter for other use-cases. For example, if a use-case requires and has the budget for large volumes of annotated data, then $VOL$ thresholds around 500-700 would be optimal. However, to retain a smaller percentage of frames, $VOL$ threshold of 800 and higher should be preferred. 
\begin{table}[htbp]
\caption{$VOL$ Filter Performance for Retaining Video Sequence Frames. Percentage of retained frames per sequence after thresolding frames for $VOL>=v$ are given.}
\begin{adjustbox}{width=\columnwidth,center}
\begin{tabular}{|c|c|c|c|c|c|c|c|}
\hline
\textbf{Video}&\multicolumn{7}{|c|}{\textbf{Varying threshold values for $v$}} \\ \cline{2-8}
\textbf{Sequence ID} & \textbf{900} & \textbf{800} & \textbf{700} & \textbf{600} & \textbf{500} & \textbf{400} & \textbf{300} \\ \hline
0000 & 26.6 & 33.1 & 48.1 & 55.2 & 62.3 & 83.8 & 100.0 \\ \hline
0001 & 16.1 & 34.0 & 49.9 & 63.3 & 74.0 & 89.7 & 100.0 \\ \hline
0002 & 17.2 & 19.3 & 32.2 & 84.1 & 100.0 & 100.0 & 100.0 \\ \hline
0003 & 67.4 & 84.0 & 94.4 & 100.0 & 100.0 & 100.0 & 100.0 \\ \hline
0004 & 0.6 & 2.2 & 13.1 & 25.5 & 61.5 & 97.5 & 100.0 \\ \hline
0005 & 0.0 & 0.0 & 0.0 & 0.0 & 10.1 & 70.0 & 100.0 \\ \hline
0006 & 0.0 & 0.0 & 0.0 & 1.1 & 86.3 & 97.0 & 100.0 \\ \hline
0007 & 48.4 & 65.6 & 75.8 & 82.3 & 90.3 & 97.9 & 100.0 \\ \hline
0008 & 0.0 & 0.5 & 3.3 & 20.8 & 66.4 & 96.9 & 100.0 \\ \hline
0009 & 9.7 & 15.3 & 24.7 & 44.1 & 70.4 & 94.5 & 99.5 \\ \hline
0010 & 34.4 & 45.9 & 60.2 & 66.7 & 93.9 & 100.0 & 100.0 \\ \hline
0011 & 0.0 & 31.6 & 50.9 & 76.9 & 90.9 & 100.0 & 100.0 \\ \hline
0012 & 0.0 & 0.0 & 0.0 & 0.0 & 100.0 & 100.0 & 100.0 \\ \hline
0013 & 3.2 & 15.3 & 37.9 & 59.1 & 99.4 & 100.0 & 100.0 \\ \hline
0014 & 41.5 & 58.5 & 86.8 & 100.0 & 100.0 & 100.0 & 100.0 \\ \hline
0015 & 0.8 & 7.2 & 15.2 & 23.4 & 100.0 & 100.0 & 100.0 \\ \hline
0016 & 0.0 & 39.7 & 96.7 & 100.0 & 100.0 & 100.0 & 100.0 \\ \hline
0017 & 53.1 & 100.0 & 100.0 & 100.0 & 100.0 & 100.0 & 100.0 \\ \hline
0018 & 91.4 & 99.1 & 100.0 & 100.0 & 100.0 & 100.0 & 100.0 \\ \hline
0019 & 12.0 & 20.4 & 31.3 & 49.6 & 81.2 & 97.4 & 100.0 \\ \hline
0020 & 0.0 & 0.0 & 0.0 & 0.0 & 6.2 & 36.1 & 92.7 \\ \hline
%\multicolumn{8}{l}{$^{\mathrm{a}}$Sample of a Table footnote.}
\end{tabular}
\label{tab3}
\end{adjustbox}
\end{table}

Next, we analyze the impact of $SSIM$ thresholds for frame retention from video sequences. We observe that by filtering frames with $SSIM<0.4$ results in the retention of around 25\% of frames. However, selecting a low $SSIM$ threshold of 0.4 would risk removing frames with novel objects not captured in prior frames. Thus, retention of frames with $SSIM<=0.7$ would be recommended for other use-cases. Table \ref{tab4} represents the percentage of retained frames per sequence by thresholding frames for $SSIM<=s$.  

\begin{table}[htbp]
\caption{$SSIM$ Filter Performance for Retaining Video Sequence Frames. Percentage of retained frames per sequence after thresolding frames for $SSIM<=s$ are given.}
\begin{adjustbox}{width=\columnwidth,center}
\begin{tabular}{|c|c|c|c|c|c|c|c|}
\hline
\textbf{Video}&\multicolumn{7}{|c|}{Varying threshold values for $s$} \\ \cline{2-8}
\textbf{Sequence} & \textbf{0.2} & \textbf{0.3} & \textbf{0.4} & \textbf{0.5} & \textbf{0.6} & \textbf{0.7} & \textbf{0.8} \\ \hline

0000 & 0.6 & 11.7 & 68.8 & 94.2 & 100.0 & 100.0 & 100.0 \\ \hline
0001 & 0.2 & 9.2 & 34.7 & 74.3 & 86.6 & 88.8 & 89.7 \\ \hline
0002 & 0.4 & 11.2 & 18.9 & 34.8 & 49.4 & 54.9 & 63.9 \\ \hline
0003 & 17.4 & 53.5 & 97.2 & 100.0 & 100.0 & 100.0 & 100.0 \\ \hline
0004 & 1.0 & 8.0 & 36.3 & 80.3 & 100.0 & 100.0 & 100.0 \\ \hline
0005 & 0.3 & 0.3 & 0.3 & 25.3 & 98.7 & 100.0 & 100.0 \\ \hline
0006 & 0.4 & 5.9 & 19.3 & 23.7 & 29.3 & 31.1 & 31.5 \\ \hline
0007 & 1.6 & 27.4 & 73.4 & 96.8 & 98.5 & 99.3 & 99.9 \\ \hline
0008 & 0.3 & 0.3 & 0.3 & 21.3 & 85.9 & 100.0 & 100.0 \\ \hline
0009 & 0.1 & 0.5 & 39.2 & 83.6 & 92.0 & 98.8 & 100.0 \\ \hline
0010 & 2.4 & 17.3 & 53.7 & 68.7 & 94.9 & 100.0 & 100.0 \\ \hline
0011 & 0.3 & 0.3 & 14.5 & 53.6 & 84.7 & 90.9 & 92.8 \\ \hline
0012 & 1.3 & 1.3 & 1.3 & 1.3 & 1.3 & 1.3 & 1.3 \\ \hline
0013 & 0.3 & 0.3 & 42.1 & 88.2 & 100.0 & 100.0 & 100.0 \\ \hline
0014 & 0.9 & 7.5 & 35.8 & 57.5 & 88.7 & 91.5 & 95.3 \\ \hline
0015 & 0.3 & 0.3 & 5.6 & 17.8 & 21.0 & 23.1 & 24.2 \\ \hline
0016 & 0.5 & 0.5 & 0.5 & 0.5 & 0.5 & 0.5 & 0.5 \\ \hline
0017 & 0.7 & 0.7 & 0.7 & 0.7 & 0.7 & 0.7 & 0.7 \\ \hline
0018 & 0.3 & 0.3 & 20.6 & 67.6 & 94.7 & 96.8 & 98.5 \\ \hline
0019 & 0.1 & 1.3 & 9.4 & 46.7 & 70.2 & 81.2 & 84.2 \\ \hline
0020 & 0.1 & 0.1 & 0.1 & 1.2 & 21.9 & 69.1 & 94.3 \\ \hline
%\multicolumn{8}{l}{$^{\mathrm{a}}$Sample of a Table footnote.}
\end{tabular}
\label{tab4}
\end{adjustbox}
\end{table}

From Tables \ref{tab3} and \ref{tab4}, we observe that filtering raw video frames for image quality first using the $VOL$ metric, followed by $SSIM$ based filtering of frames to exclude images with similar content can be modified to suit the use case. Altering the order of filtering or parallelizing the two processes may lead to uneven quality in retained/training data. Also, $SSIM$ is highly dependent on the frame rate of the video sequence. For higher frame-rate video sequences, several consecutive frames may appear similar, while for lower frame rate the frame content can change drastically, with relatively lower $SSIM$ metrics. Thus, for filtering frames from video sequences with high frame rate, an average $VOL$ threshold (around 500) and $SSIM$ around 0.7 may be significant. However, for filtering frames form a low frame rate video, higher $VOL$ thresholds (around 800) and lower $SSIM$ thresholds (around 0.2) may be sufficient. 

\subsection{Video-sequence Classification}
Upon retaining a subset of video frames for training purpose, we further generate auto-tagged metadata for frames per sequence using rule-based aggregations over the outcome of object detectors. The performance of the proposed automated scene categorization operation for the various video sequences is shown in Table \ref{tab5}. Also, the performance of the automated scene categorization is compared to manually generated scene categories. The final automatically selected scene category represents the category with the maximum number of frames in the sequence. In Table \ref{tab5}, the highest percentage of frame categories are in bold and whenever the automated category agrees with the manual categorization, the manual categorization entry is in bold.
\begin{table*}[htbp]
\caption{Scene Categorization Performance for automated tagging of filtered frames.}
\begin{center}
\begin{tabular}{|c|c|c|c|c|c|c|}
\hline
\textbf{Video} & \textbf{Frame} &\multicolumn{4}{|c|}{\textbf{Percentage of Frames}} & \textbf{Manual}\\
\cline{3-6} 
\textbf{Sequence} & \textbf{Count} & \textbf{City}  & \textbf{Rural}  & \textbf{Freeway}  & \textbf{Pedestrian} & \textbf{Categorization} \\ \hline
0000 & 154 & \textbf{96.75} & 3.25 & 0.00 & 0.00 & {\bf City} \\ \hline
0001 & 447 & 17.45 & 8.28 & \textbf{74.27} & 0.00 & City \\ \hline
0002 & 233 & \textbf{77.25} & 11.16 & 11.59 & 0.00 & Freeway \\ \hline
0003 & 144 & 0.00 & 13.89 & \textbf{86.11} & 0.00 & {\bf Freeway} \\ \hline
0004 & 314 & 28.34 & 11.15 & \textbf{60.51} & 0.00 & {\bf Freeway} \\ \hline
0005 & 297 & 46.80 & 1.68 & \textbf{51.52} & 0.00 & {\bf Freeway} \\ \hline
0006 & 270 & 0.00 & 31.85 & \textbf{68.15} & 0.00 & {\bf Freeway} \\ \hline
0007 & 800 & 5.50 & 37.75 & \textbf{56.13} & 0.63 & City \\ \hline
0008 & 390 & 0.00 & 5.38 & \textbf{94.62} & 0.00 & {\bf Freeway} \\ \hline
0009 & 803 & 3.61 & 38.61 & \textbf{57.78} & 0.00 & {\bf Freeway} \\ \hline
0010 & 294 & 25.85 & 32.31 & \textbf{41.84} & 0.00 & {\bf Freeway} \\ \hline
0011 & 373 & 15.28 & 4.56 & \textbf{80.16} & 0.00 & City \\ \hline
0012 & 78 & \textbf{84.62} & 15.38 & 0.00 & 0.00 & {\bf City} \\ \hline
0013 & 340 & 13.53 & 12.06 & 0.00 & \textbf{67.65} & {\bf Pedestrian} \\ \hline
0014 & 106 & \textbf{54.72} & 2.83 & 42.45 & 0.00 & {\bf City} \\ \hline
0015 & 376 & \textbf{92.82} & 6.65 & 0.00 & 0.53 & {\bf City} \\ \hline
0016 & 209 & \textbf{100.00} & 0.00 & 0.00 & 0.00 & {\bf City} \\ \hline
0017 & 145 & 0.00 & 0.00 & 0.00 & \textbf{100.00} & {\bf Pedestrian} \\ \hline
0018 & 339 & 0.00 & 15.93 & \textbf{84.07} & 0.00 & City \\ \hline
0019 & 1059 & \textbf{50.05} & 7.84 & 0.19 & 8.50 & Pedestrian \\ \hline
0020 & 837 & 0.00 & 0.00 & \textbf{100.00} & 0.00 & {\bf Freeway} \\ \hline
\end{tabular}
\label{tab5}
\end{center}
\end{table*}
From Table \ref{tab5} We observe that several `City' sequences such as `0001', `0007', `0011', and `0018' are incorrectly categorized by the automated system as `Freeway' sequences. Upon investigation, we observe that many of the vehicles in such sequences were parked by the edges of the road in the city scenes. This increases the overall count of vehicles in a scene, thereby appearing as a `Freeway' scene to an automated system that typically contains several running vehicles. A possible solution and future work to resolve this issue would be to include more context into scene categorization, such as relative object location. Thus, detecting vehicles along with a pavement/road edge should signify a 'City' scene as opposed to a `Freeway'. Also, detection of buildings and relative distance of vehicles to the buildings could be alternative methods to correctly classify these sequences as `City'.

Also, in Table \ref{tab5} we observe that for the `0002' sequence, that is a `Freeway' sequence, the automated system misclassifies it as a `City' sequence. This is because there are retained frames in this sequence from a large intersection between two Freeway roads with a pedestrian standing at a crosswalk. Presence of the pedestrian signifies a `City' scene for the automated system. To resolve this classification, detection of more contextual objects such as freeway landmarks, road divider etc. can be beneficial.

Additionally, from Table \ref{tab5} we observe that the sequence `0019', which is a `Pedestrian' sequence gets automatically mis-classified as a `City' sequence. Further investigation shows that this sequence is an ambiguous sequence, with several vehicles and some pedestrians in few frames. The automated system gives more importance to vehicle detection than to pedestrians detection thereby resulting in a mis-classification. To resolve this issue, objects can be given different weight ages (e.g. `Pedestrians' can get a higher weightage than `Vehicles'). Also, this sequence is an outlier that requires more manual supervision for tagging than the others. 

\begin{figure}[ht]
	\includegraphics[width=0.45\textwidth]{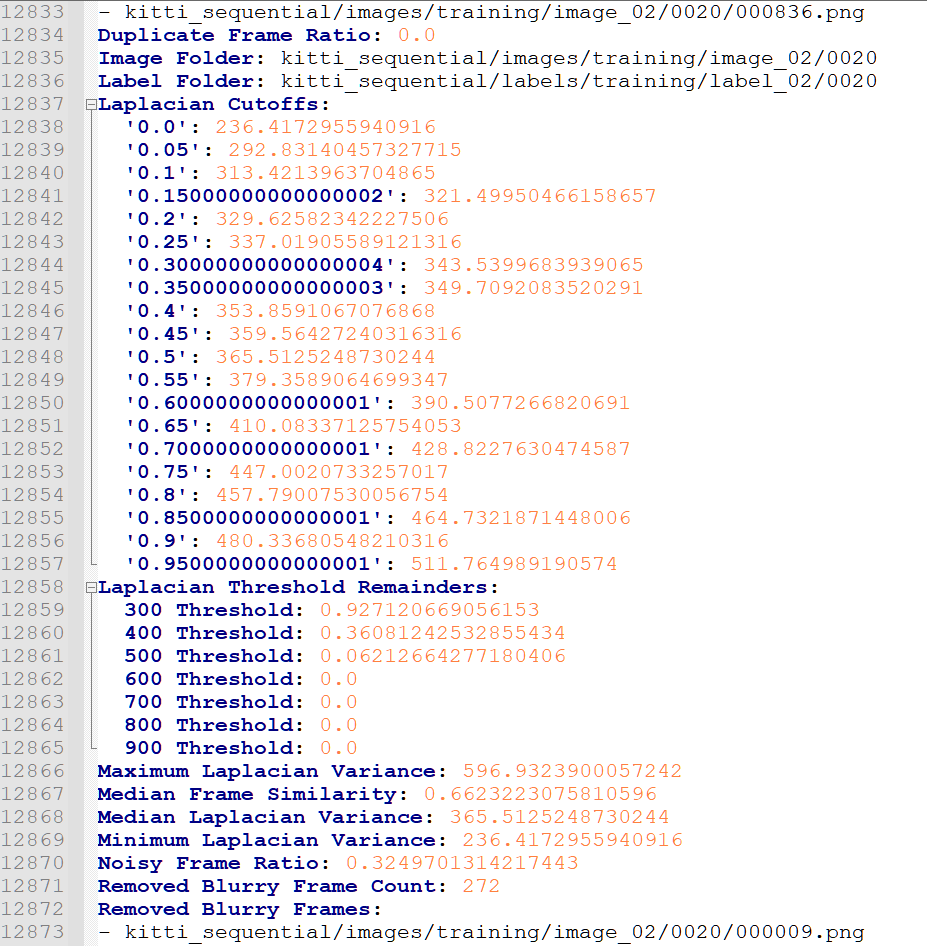}
	\caption{Example yaml output file for sequence 0020.}\label{ex4}
\end{figure}

Finally, a dictionary of retained and removed frames along with the scene categorization is stored in yaml format for storage and retrieval purposes as shown in Fig. \ref{ex4}.

For overall system-level considerations, the processing time per sequence is also recorded in the yaml file exports. Analysis of all 22 sequences from the KITTI data set resulted in around 100 frames being retained in about 29 seconds. This includes verifying a deployed ML model for object detection performance on the filtered/test images. Thus, building the proposed pipeline will lead to consistently stored and tagged video frames, that can significantly reduce delays in ML and deployment pipelines.

\subsection{Performance Analysis on Non-sequential Image Data}
In addition to analyzing the performance of the proposed video-data pipeline on video sequences, we analyze its impact on isolating \textit{high quality} and \textit{high content diversity} frames in non-sequential data batches. Here, we assess the performance of $VOL$ and $SSIM$ thresholds on the KITTI 2D Object Detection dataset. This dataset contains outdoor images that are all out of sequence. Here, we observe that a $VOL>900$ threshold reduces the dataset to 22.6\% of the total original frames, while $SSIM<0.2$ threshold eliminates similar scenery structures and reduces the dataset to 64\% of the frames. The performance of frame reduction at varying thresholds is given in Table \ref{tab6}. Thus, for non-sequential datasets, using only $VOL$ for frame filtering can be sufficient, followed by object detection per frame and frame categorization prior to storage.
\begin{table}[htbp]
\caption{Performance of $VOL$ and $SSIM$ for frame reduction in non-sequential data. Percentage of the retained frames is shown here.}
\begin{center}
\begin{tabular}{|c|c|c|c|c|c|c|}
\hline
\cline{1-7} 

\multicolumn{7}{|c|}{\textbf{$VOL$ threshold values}}  \\ \hline
\textbf{900} & \textbf{800} & \textbf{700} & \textbf{600} & \textbf{500} & \textbf{400} & \textbf{300} \\ \hline
22.6 & 36.0 & 47.4 & 59.3 & 79.7 & 93.1 & 99.6 \\ \hline
\multicolumn{7}{|c|}{\textbf{$SSIM$ threshold values}} \\ \hline
\textbf{0.2} & \textbf{0.3} & \textbf{0.4} & \textbf{0.5} & \textbf{0.6} & \textbf{0.7} & \textbf{0.8} \\ \hline
63.9 & 96.3 & 99.4 & 99.7 & 99.7 & 99.8 & 99.9 \\ \hline
\end{tabular}
\label{tab6}
\end{center}
\end{table}

Further analysis of the fine-tuned and deployed ML models resulted in 70\% average precision across test frames for detecting `Vehicles' and 58\% average precision for detecting `People' using the FasterRCNN model. The average precision for fine-tuned Mask-RCNN were 72\% and 62\% for `Vehicles' and `People', respectively.

%%%%%%%%%%%%%%%%%%%%%%
%%%%%%%%%%%%%%%%%%%%%%%%%%%%%%%%%%%%%%%%%%%%%%%%%%%%%%%%%

\section{Conclusions and Discussion}
In this work we present a two-step video-data pipeline that is capable of reviewing sequential stack of frames to retain images with high quality and high variations in content. Such data pipelines further aid ML modeling and deployment pipelines. We analyze the importance of various thresholds applied in the data pipeline towards controlling for the ratio of retained over removed frames per sequence. Additionally, we assess the applicability for the proposed video-data pipeline for varying use-cases that include video sequences with varying frame rates and non-sequential image frames we well.

In addition, the proposed video-data pipeline enables sequence metadata generation and tagging for retained frames that becomes useful for automated training data collection for model updates. We apply the proposed video-data pipeline to train multiple versions of ML models that can be rolled out to delivery over a Kubernetes cluster. We design a deployment and monitoring pipeline to automatically detect drifts in online data to enable automated versioned deployments with minimal manual supervision. Thus, the proposed video-data pipeline can be used to scale and further automate the process of taking ML solutions to production.

The proposed video-data pipeline is found to be most efficient for outdoor autonomous drive sequences. However, it can scale for indoor videos from handheld cameras as well, since the image quality detection module can effectively discard \textit{blurry} frames or frames with \textit{artifacts}. The video-data pipeline will need newer object and scene categories for such a modification.

It is noteworthy that the proposed pipeline will not scale for 3D medical image stacks since the pixel variations and structural similarities in medical images (CT, Xray) require more contrastive models for identifying variations such as in \cite{SISE-PC}. Future works may be directed towards optimizing the proposed video-data pipeline for medical image or LIDAR-data use cases. 
\bibliographystyle{mlsys2022}
\bibliography{papers}

\end{document}